\documentclass[10pt,twocolumn,letterpaper]{article}

\usepackage{iccv}
\usepackage{times}
\usepackage{epsfig}
\usepackage{graphicx}
\usepackage{amsmath}
\usepackage{amssymb}
\usepackage{multirow}
\usepackage{bbm}
\usepackage{rotating}
\usepackage{array}
\usepackage{threeparttable}
\allowdisplaybreaks[4]

\usepackage{epstopdf}
\usepackage{subfigure}
\usepackage{amsmath}
\usepackage{bm}
\usepackage{url}
\usepackage{algorithm}
\usepackage{algorithmic}

\usepackage[breaklinks=true,bookmarks=false]{hyperref}

\iccvfinalcopy 

\def\iccvPaperID{****} 

\begin{document}

\title{Racial Faces in-the-Wild: Reducing Racial Bias by Information Maximization Adaptation Network}

\author{Mei Wang\textsuperscript{1}, Weihong Deng\textsuperscript{1*}, Jiani Hu\textsuperscript{1}, Xunqiang Tao\textsuperscript{2}, Yaohai Huang\textsuperscript{2}\\
\textsuperscript{1}Beijing University of Posts and Telecommunications, \textsuperscript{2}Canon Information Technology (Beijing) Co., Ltd\\
{\tt\small \textsuperscript{1}\{wangmei1, whdeng, jnhu\}@bupt.edu.cn, \textsuperscript{2}\{taoxunqiang, huangyaohai\}@canon-ib.com.cn}}

\maketitle

\begin{abstract}
Racial bias is an important issue in biometric, but has not been thoroughly studied in deep face recognition. In this paper, we first contribute a dedicated dataset called Racial Faces in-the-Wild (RFW) database, on which we firmly validated the racial bias of four commercial APIs and four state-of-the-art (SOTA) algorithms. Then, we further present the solution using deep unsupervised domain adaptation and propose a deep information maximization adaptation network (IMAN) to alleviate this bias by using Caucasian as source domain and other races as target domains. This unsupervised method simultaneously aligns global distribution to decrease race gap at domain-level, and learns the discriminative target representations at cluster level. A novel mutual information loss is proposed to further enhance the discriminative ability of network output without label information. Extensive experiments on RFW, GBU, and IJB-A databases show that IMAN successfully learns features that generalize well across different races and across different databases.

\end{abstract}

\section{Introduction}

The emergence of deep convolutional neural networks (CNN) \cite{krizhevsky2012imagenet,simonyan2014very,szegedy2015going,he2016deep,hu2017squeeze} greatly advances the frontier of face recognition (FR) \cite{wang2018deep,sun2014deep,schroff2015facenet}. 
However, more and more people find that a problematic issue, namely racial bias, has always been concealed in the previous studies due to biased benchmarks but it explicitly degrades the performance in realistic FR systems \cite{MITREVIEW,pmlr-v81-buolamwini18a,garvie2016perpetual,alvi2018turning}. For example, Amazon's Rekognition Tool incorrectly matched the photos of 28 U.S. congressmen with the faces of criminals, especially the error rate was up to 39\% for non-Caucasian people. 
Although several studies \cite{phillips2003face,grother2010report,furl2002face,phillips2011other,klare2012face} have uncovered racial bias in non-deep FR algorithms, this field still remains to be vacant in deep learning era because so little testing information available makes it hard to measure the racial bias.

\begin{figure}[htbp]
\centering
\includegraphics[width=8cm]{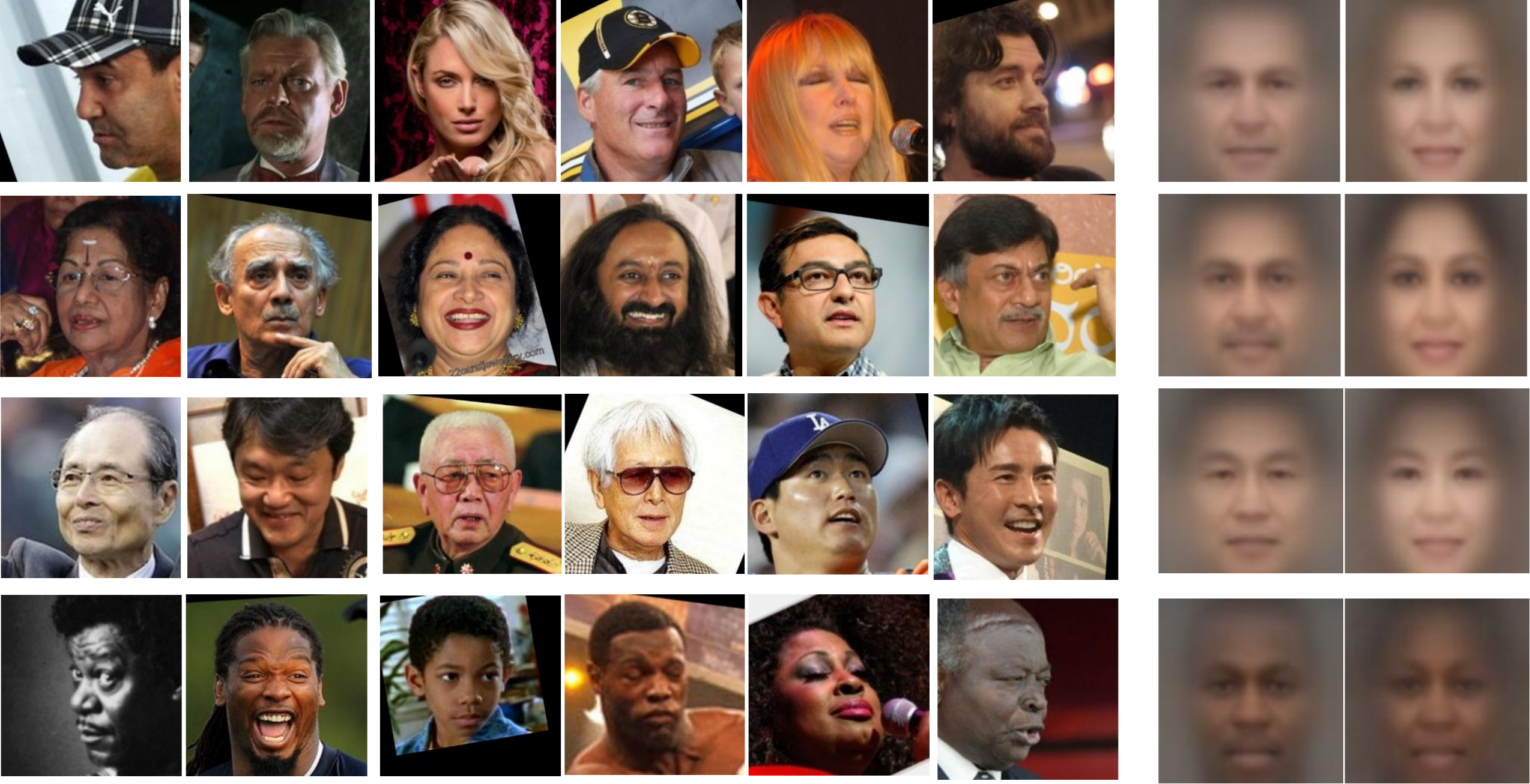}
\caption{ Examples and average faces of RFW database. In rows top to bottom: Caucasian, Indian, Asian, African. }
\label{fig1}
\end{figure}

To facilitate the research towards this issue, in this work we construct a new Racial Faces in-the-Wild (RFW) database, as shown in Fig. \ref{fig1} and Table \ref{tab4}, to fairly measure racial bias in deep FR. Based on experiments on RFW, we find that both commercial APIs and SOTA algorithms indeed suffer from racial bias: the error rates on African faces are about two times of Caucasians, as shown in Table \ref{tab5}. To investigate the biases caused by training data, we also collect a race-balanced training database, and validate that racial bias comes on both data and algorithm aspects. Some specific races are inherently more difficult to recognize even trained on the race-balanced training data. Further research efforts on algorithms are requested to eliminate racial bias.

\begin{table}[htbp]
    \footnotesize
	\begin{center}
    \begin{threeparttable}
    \setlength{\tabcolsep}{1.5mm}{
	\begin{tabular}{c|c|cccc}
		\hline
        & \multirow{2}{*}{Model}  & \multicolumn{4}{c}{RFW} \\
        &                                            & Caucasian & Indian & Asian & African \\ \hline \hline
        &  Microsoft \cite{azure}  &87.60 &82.83 &79.67 &75.83   \\
        commercial &  Face++ \cite{Face++} & 93.90& 88.55& 92.47& 87.50\\
        API &  Baidu \cite{Baidu} &89.13 & 86.53& 90.27&77.97  \\
        &  Amazon \cite{amazon}  &90.45 &87.20 &84.87 &86.27  \\ \cline{2-6}
        &  mean  & 90.27 & 86.28 & 86.82 & 81.89 \\ \hline
        &  Center-loss \cite{wen2016discriminative}  & 87.18 &81.92 & 79.32&78.00 \\
        SOTA &  Sphereface \cite{liu2017sphereface} & 90.80 &87.02 &82.95 &82.28  \\
        algorithm & Arcface\tnote{1} \cite{deng2018arcface}  & 92.15 &88.00 &83.98 &84.93  \\
        & VGGface2 \cite{cao2017vggface2}  &89.90 &86.13 &84.93 &83.38  \\ \cline{2-6}
        & mean  & 90.01 & 85.77 & 82.80 & 82.15 \\ \hline
	\end{tabular}}
    \begin{tablenotes}
     \item[1] Arcface here is trained on CASIA-Webface using ResNet-34. 
    \end{tablenotes}
    \end{threeparttable}
    \end{center}
    \caption{Racial bias in deep FR systems. Verification accuracies (\%) evaluated on 6000 difficult pairs of RFW database are given. }
    \label{tab5}
\end{table}

Unsupervised domain adaptation (UDA) \cite{wang2018deepdomain} is one of the promising methodologies to address algorithm biases, which can map two domains into a domain-invariant feature space and improve target performances in an unsupervised manner \cite{Tzeng2014Deep,Long2015Learning,
Tzeng2017Adversarial,Ganin2015Unsupervised}. Unfortunately, most UDA methods for object recognition are not applicable for FR because of two unique challenges. First, face identities (classes) of two domains are non-overlapping in FR, so that many skills in state-of-the-art (SOTA) methods based on sharing classes are inapplicable. Second, popular methods by the global alignment of source and target domain are insufficient to acquire the discriminating power for classification in FR. How to meet these two challenges is meaningful but few works have been proposed in this community.

In this paper, we propose a new information maximization adaptation network (IMAN) to mitigate racial bias, which matches global distribution at domain-level, at the meantime, learns discriminative target distribution at cluster-level. To circumvent the non-overlapping classes between two domains, IMAN applies a spectral clustering algorithm to generate pseudo-labels, by which the network is pre-adapted with Softmax and the target performance is enhanced preliminarily. This clustering scheme of IMAN is fundamentally different from other UDA methods  \cite{saito2017asymmetric,zhang2018collaborative,chen2018progressive,chen2011co} that are inapplicable to FR. Besides pseudo label based pre-adaptation, a novel mutual information (MI) based adaptation is proposed to further enhance the discriminative ability of the network output, which learns larger decision margins in an unsupervised way. Different from the common supervised losses and supervised MI methods \cite{singh2018supervised,jun2011compact}, MI loss takes advantage of all unlabeled target data, no matter whether they are successfully assigned pseudo-labels or not, in virtue of its unsupervised property.

Extensive experimental results show that IMAN conducted to transfer recognition knowledge from Caucasian (source) domain to other-race (target) domains. Its performance is much better than other UDA methods. Ablation study shows that MI loss has unique effect on reducing racial bias. In addition, IMAN is also helpful in adapting general deep model to a specific database, and achieved improved performance on GBU \cite{Phillips2012The} and IJB-A \cite{klare2015pushing} databases. The contributions of this work are three aspects. 1) A new RFW dataset is constructed and is released \footnote{http://www.whdeng.cn/RFW/index.html} for the study on racial bias. 2) Comprehensive experiments on RFW validate the existence and cause of racial bias in deep FR algorithms. 3) A novel IMAN solution is introduced to address racial bias.

\section{Related work}

\textbf{Racial bias in face recognition.} Several studies \cite{phillips2003face,grother2010report,furl2002face,phillips2011other,klare2012face} have uncovered racial bias in non-deep face recognition algorithms. The FRVT 2002 \cite{phillips2003face} showed that recognition accuracies depend on demographic cohort. 
Phillips et al. \cite{phillips2011other} evaluated FR algorithms on the images of FRVT 2006 \cite{beveridge2008focus} and found that algorithms performed better on natives. Klare et al. \cite{klare2012face} collected mug shot face images of White, Black and Hispanic from the Pinellas County Sheriff's Office (PCSO) and concluded that the Black cohorts are more difficult to recognize. In deep learning era, existing racial bias databases are no longer suitable for deep FR algorithms due to their small scale and constrained conditions; commonly-used testing databases of deep FR, e.g. LFW \cite{huang2007labeled}, IJB-A \cite{klare2015pushing}, don't include significant racial diversity, as shown in Table \ref{tab1}. Although some studies, e.g. unequal-training \cite{aminiuncovering} and suppressing attributes \cite{alvi2018turning,mirjalili2018gender,othman2014privacy,mirjalili2018semi}, have made effort to mitigate racial and gender bias in several computer vision tasks, this study remains to be vacant in FR.  
Thus, we construct a new RFW database to facilitate the research towards this issue.

\begin{table}[htbp]
	\begin{center}
    \footnotesize
    \setlength{\tabcolsep}{1.5mm}{
	\begin{tabular}{c|c|cccc}
		\hline
       Train/ & \multirow{2}{*}{Database} & \multicolumn{4}{c}{Racial distribution (\%)} \\
       Test& & Caucasian & Asian & Indian & African\\ \hline \hline
       \multirow{3}{*}{train} & CASIA-WebFace \cite{yi2014learning} & 84.5&2.6 &1.6 &11.3  \\
                                 & VGGFace2 \cite{cao2017vggface2} &74.2 &6.0 & 4.0&15.8 \\
                                 & MS-Celeb-1M \cite{guo2016ms} &76.3&6.6 &2.6 &14.5 \\ \hline
        \multirow{3}{*}{test} & LFW \cite{huang2007labeled} &69.9 & 13.2& 2.9& 14.0 \\
                                 & IJB-A \cite{klare2015pushing} & 66.0& 9.8& 7.2& 17.0 \\
                                 & RFW & \textbf{25.0}& \textbf{25.0}& \textbf{25.0}& \textbf{25.0}\\ \hline
	\end{tabular}}
    \end{center}
    \caption{The percentage of different race in commonly-used training and testing databases}
    \label{tab1}
\end{table}



\textbf{Deep unsupervised domain adaptation.} UDA \cite{wang2018deepdomain} utilizes labeled data in relevant source domains to execute new tasks in a target domain \cite{Tzeng2014Deep,Long2015Learning,Long2016Deep,Ganin2015Unsupervised,Tzeng2017Adversarial}. However, the research of UDA is limited to object classification, very few studies have focused on UDA for FR task. Luo et al. \cite{Luo2018Adaptation} integrated the maximum mean discrepancies (MMD) estimator to CNN to decrease domain discrepancy. Sohn et al. \cite{sohn2017unsupervised} synthesized video frames from images by a set of transformations and applied a domain adversarial discriminator to align feature space of image and video domains. 
Kan et al. \cite{kan2015bi} utilized the sparse representation constraint to ensure that source domain shares similar distribution as target domain. 
In this paper, inspired by Inception Score \cite{salimans2016improved,barratt2018note} used in Generative Adversarial Nets (GAN), we introduce MI as a regularization term to domain adaptation and propose a novel IMAN method to address this unique challenge of FR in an unsupervised way.

\section{Racial Faces in-the-Wild: RFW}


Instead of downloading images from websites, we collect them from MS-Celeb-1M \cite{MSCH3}. We use the ``Nationality'' attribute of FreeBase celebrities \cite{freebase} to directly select Asians and Indians. For Caucasians and Africans, Face++ API \cite{Face++} is used to estimate race. An identity will be accepted only if its most images are estimated as the same race, otherwise it will be abandoned. To avoid the negative effects caused by the biased Face++ tool, we manually check some images with low confidence scores from Face++.

Then we construct our RFW database with four testing subsets, namely Caucasian, Asian, Indian and African. Each subset contains about 10K images of 3K individuals for face verification. All of these images have been carefully and manually cleaned. Besides, in order to exclude overlapping identities between RFW and commonly-used training datasets, we further remove the overlapping subjects by manual inspection, when the subject and its nearest neighbor in CASIA-Webface and VGGFace2 (based on Arcface \cite{deng2018arcface} feature) are found to be of the same identity.

For the performance evaluation, we recommend to use both the biometric receiver operating characteristic (ROC) curve and LFW-like protocol. Specifically, ROC curve, which aims to report a comprehensive performance, evaluates algorithms on all pairs of 3K identities (about 14K positive vs. 50M negative pairs). In contrast, LFW-like protocol facilitates easy and fast comparison between algorithms
with 6K pairs of images. Further, inspired by the ugly subset of GBU database \cite{Phillips2012The}, we have selected the ``difficult" pairs (in term of cosine similarity) to avoid the saturated performance to be easily reported \footnote{All data and baseline code for evaluating will be publicly available for the research purpose.}.

\begin{figure}[htbp]
\centering
\includegraphics[width=7cm]{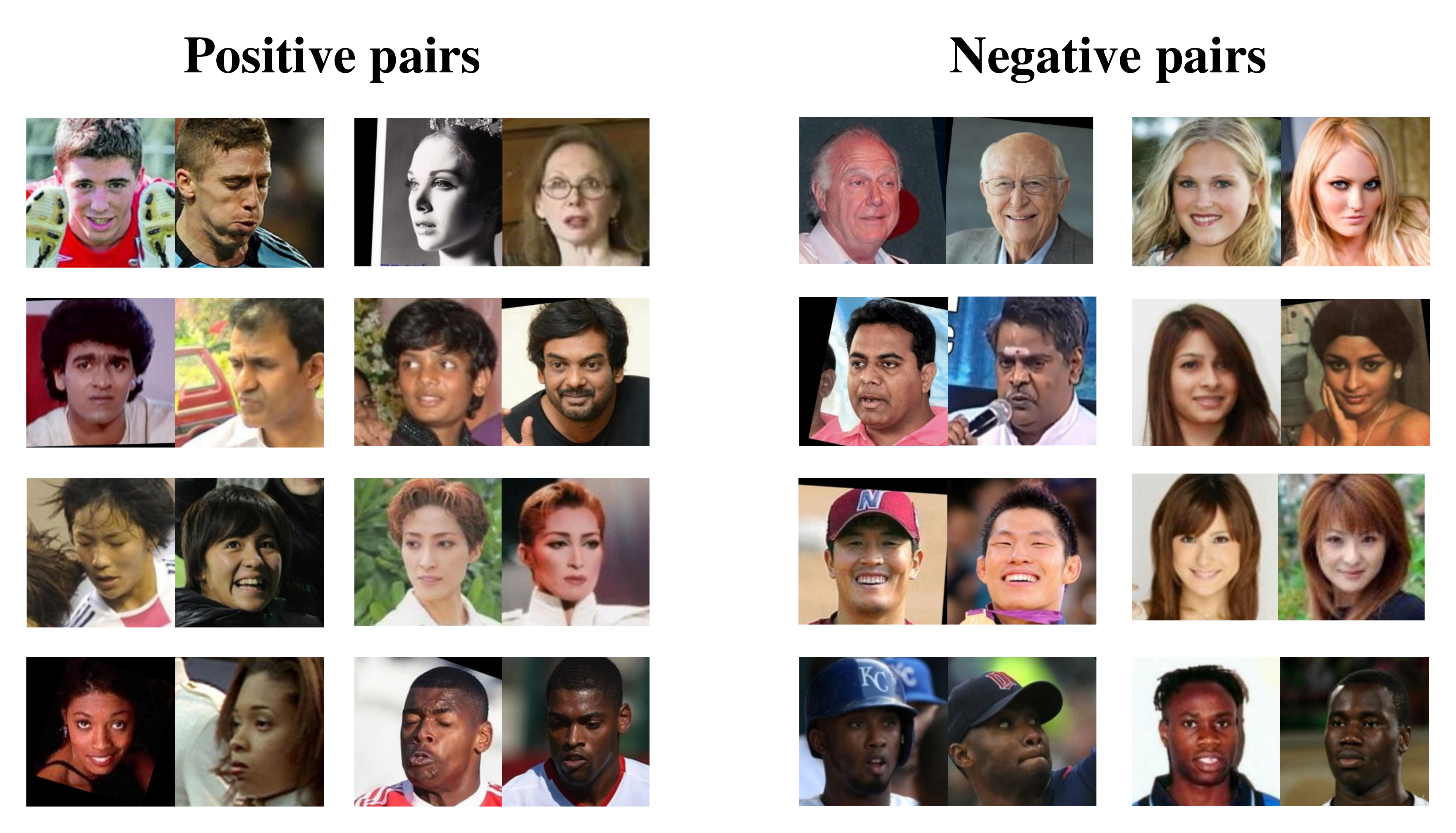}
\caption{ Examples of pairs in RFW database. We select 6K difficult pairs according to cosine similarity to avoid saturated performance, these images challenge the recognizer by variations of same people and the similar appearance of different people.}
\label{fig8}
\end{figure}

\begin{figure*}[htbp]
\centering
\subfigure[yaw]{
\label{fig10a} 
\includegraphics[width=2.6cm]{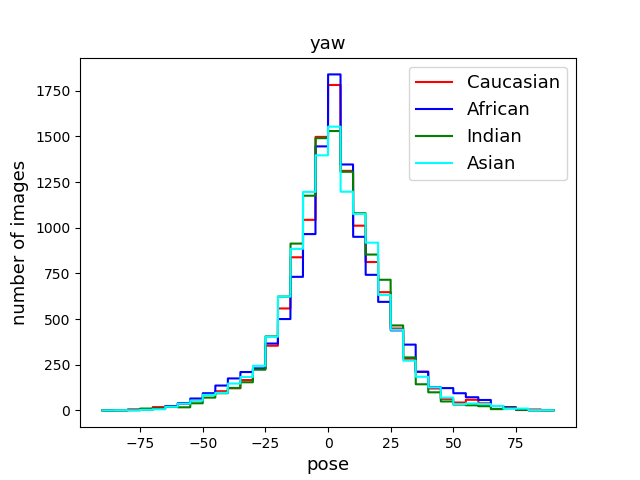}}
\hspace{0cm}
\subfigure[pitch]{
\label{fig10b} 
\includegraphics[width=2.6cm]{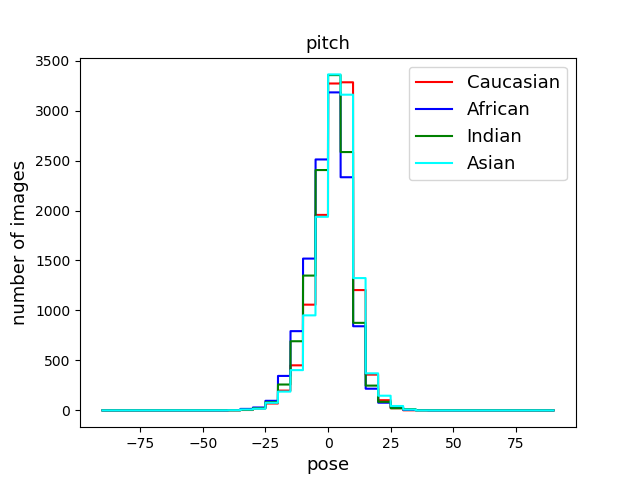}}
\hspace{0cm}
\subfigure[age]{
\label{fig10c} 
\includegraphics[width=2.6cm]{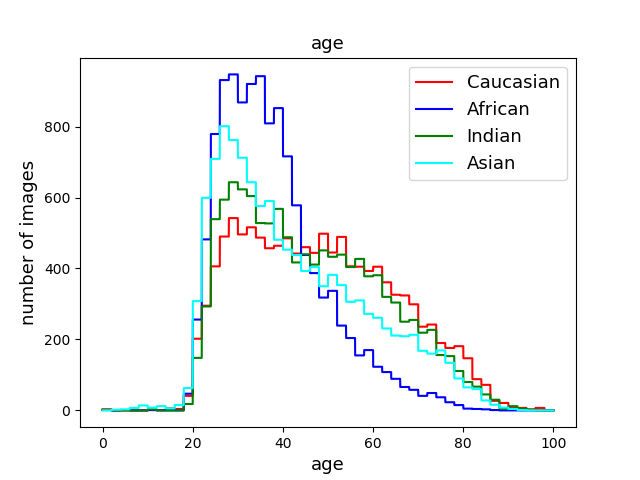}}
\hspace{0cm}
\subfigure[gender]{
\label{fig10d} 
\includegraphics[width=2.6cm]{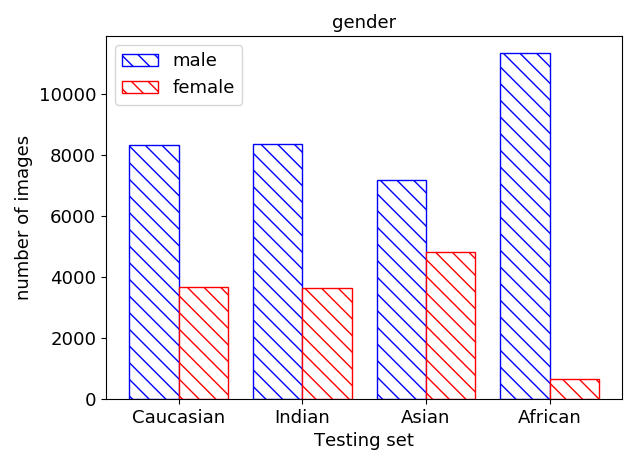}}
\hspace{0cm}
\subfigure[pose gap]{
\label{fig10e} 
\includegraphics[width=2.6cm]{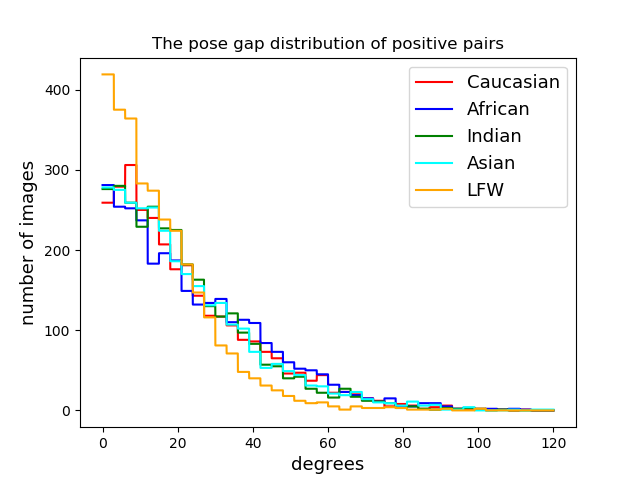}}
\hspace{0cm}
\subfigure[age gap]{
\label{fig10f} 
\includegraphics[width=2.6cm]{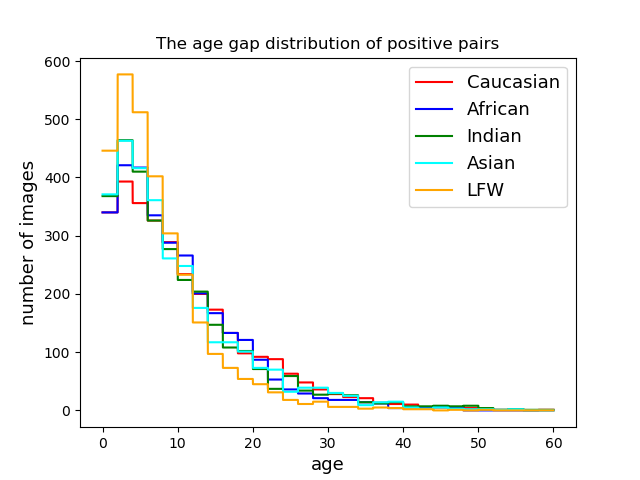}}
\caption{ RFW statistics. We show the (a) yaw pose, (b) pitch pose, (c) age and (d) gender distribution of 3000 identities in RFW, as well as (e) Pose gap distribution and (f) age gap distribution of positive pairs in LFW and RFW.}
\label{fig10} 
\end{figure*}

In RFW, the images of each race are randomly collected from MS-Celeb-1M without any preference, and thus they are suitable to fairly measure racial bias. We have validated that, across varying races, their distributions of pose, age, and gender are similar. As evidence, the detailed distributions measured by Face++ API are show in Fig. \ref{fig10a}-\ref{fig10d}. One can see from the figures that there is no significant difference between different races.

Moreover, the pose and age gap distributions of 3K difficult positive pairs are show in Fig. \ref{fig10e} and \ref{fig10f}, which indicates that the selected difficult pairs are also fair across different races and contain larger intra-person variations. And Fig. \ref{fig8} presents some examples of the 6K selected pairs, and one can see from the figure that some pairs are very challenging even for human.

\section{Information maximization adaptation network}

\begin{figure*}[htbp]
\centering
\includegraphics[width=17cm]{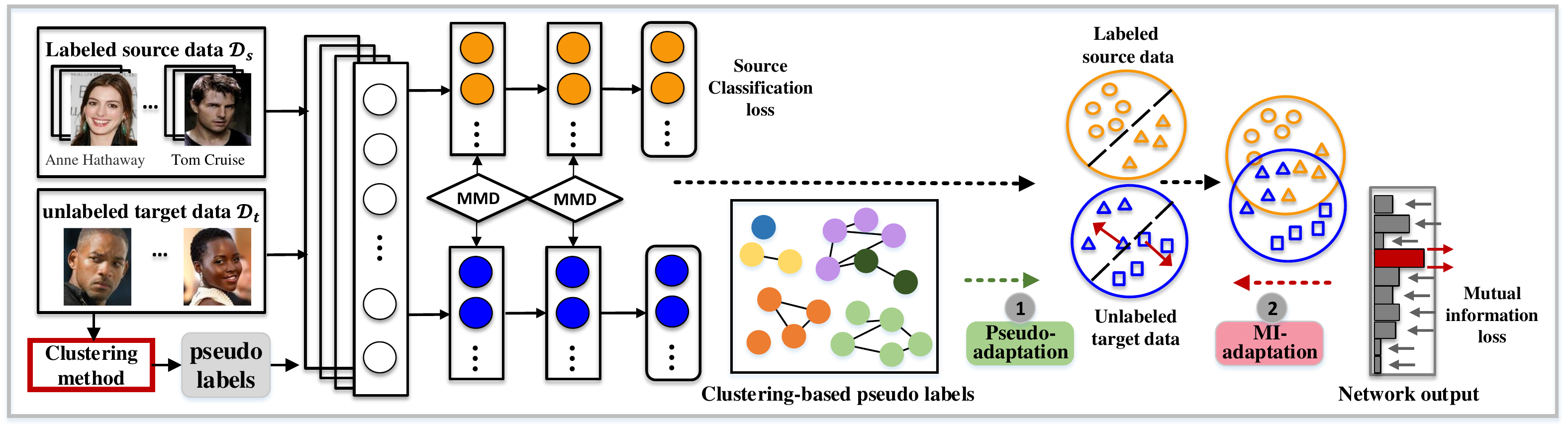}
\caption{Overview of IMAN architecture. \textbf{Step-1: Pseudo-adaptation.} Pseudo-labels of target images are generated by clustering algorithm and then are utilized to pre-adapt the network with supervision of Softmax to obtain preliminary improvement of target domain. \textbf{Step-2: MI-adaptation.} With mutual information loss, the distribution of target classifier's output is further optimized and larger decision margins are learned without any label information. }
\label{fig9} 
\end{figure*}

In our study, source domain is a labeled training set, namely $\mathcal{D}_{s}=\{{x^{s}_{i}},{y^{s}_{i}}\}^{M}_{i=1}$ where $x^{s}_{i}$ is the $i$-th source sample, $y^{s}_{i}$ is its category label, and $M$ is the number of source images. Target domain is an unlabeled training set, namely $\mathcal{D}_{t}=\{{x^{t}_{i}}\}^{N}_{i=1}$ where $x^{t}_{i}$ is the $i$-th target sample and $N$ is the number of target images. The data distributions of two domains are different, $P(X_s,Y_s)\neq P(X_t,Y_t)$. Our goal is to learn deep features invariant between domains and improve the performance of target images (faces of colored skin in our study) in an unsupervised manner. In the face recognition task, the identities (class) of two domains are non-overlapping, which poses a unique challenge different from other tasks.

\subsection{Clustering-based pseudo labels for pre-adaptation}

Previous UDA methods apply the source classifier to predict pseudo-labels in the target domain, by which the network can be fine-tuned using supervised losses \cite{saito2017asymmetric,zhang2018collaborative,chen2018progressive,chen2011co,xie2018learning}. Unfortunately, these well-established approaches are inapplicable in face recognition due to the non-overlapping identities between two domains. Therefore, we introduce a clustering algorithm into UDA to generate pseudo-labels for pre-adaptation training. The detailed steps of our clustering algorithm are given as following:

First, we feed unlabeled target data $X_{t}$ into network and extract deep features $\mathcal{F}(X_{t})$. Then, with these deep presentations, we construct a $N \times N$ adjacency matrix, where $N$ is the number of faces in target domain and entry at $(i,j)$, i.e. $s(i,j)$, is the cosine similarity between target face $x^{t}_{i}$ and $x^{t}_{j}$.

Second, we can build a clustering graph $\mathcal{G}(n,e)$ according to adjacency matrix, where the node $n_{i}$ represents $i$-th target image and edge $e(n_{i},n_{j})$ signifies that two target images have larger cosine-similarity than the parameter $\lambda$:
\begin{equation}
e(n_{i},n_{j})=\left\{
             \begin{array}{lll}
             1, & if \ s(i,j)>\lambda & \\
             0, & otherwise& \\
             \end{array}
\right. \label{cluster}
\end{equation}
Then, we simply save each connected component with at least $p$ nodes as a cluster (identity) and obtain pseudo-labels of these target images; the remaining images will be abandoned. So, we only obtain pseudo-labels of partial images with higher confidence to alleviate negative influence caused by falsely-labeled samples. After that, we pre-adapt the network with the standard Softmax loss.

\subsection{Mutual information loss for discriminant adaptation}

Although pre-adaptation has derived preliminary prediction of the target images, it is insufficient to boost the performance in target domain due to the imperfection of pseudo-labels. How can we take full advantage of the full set of target images and learn more discriminative representations? Based on the preliminary prediction, we propose to further optimize the distribution of classifier's output without any label information. Our idea is to learn large decision margins in feature space through enlarging the classifier's output of one class while suppressing those of other classes in an unsupervised way. Different from supervised mutual information \cite{singh2018supervised,chen2016infogan,pan2010cross,jun2011compact}, our MI loss maximizes mutual information between unlabeled target data $\mathbf{X_t}$ and classifier's prediction $\mathbf{O_t}$ inspired by \cite{Yuan2012Information,Gomes2010Discriminative}.

Based on the desideratum that an ideal conditional distribution of classifier's prediction $p(\mathbf{{O_t}}|x^t_i)$ should look like $[0,0,...,1,...,0]$, it's better to classify samples with large margin. Grandvalet \cite{grandvalet2005semi} proved that a entropy term $\tfrac{1}{N}\sum_{i=1}^{N}H(\mathbf{O_t}|x^t_i)$ very effectively meets this requirement, because it is maximized when the distribution of classifier's prediction is uniform and vice versa. However, in the case of fully unsupervised learning, simply minimizing this entropy will cause that more decision boundaries are removed and most samples are assigned to the same class. Therefore, we prefer to uniform distribution of category. An estimate of the marginal distribution of classifier's prediction $p(\mathbf{O_t})$ is given as follows:
\begin{equation}
p(\mathbf{O_t})=\int p(x^t_i)p(\mathbf{O_t}|x^t_i)dx^t_i=\tfrac{1}{N}\sum_{i=1}^{N} p\left ( \mathbf{O_t}|x^t_i \right )
\end{equation}
we suggest that maximizing the entropy of $\mathbf{O_t}$ can make samples assigned evenly across the categories of dataset.

In information theory, mutual information between $X$ and $Y$ , i.e. $I(X;Y)$, can be expressed as the difference of two entropy terms:
\begin{equation}
I(X;Y)= H(X)-H(X|Y) = H(Y)-H(Y|X)
\end{equation}
If $X$ and $Y$ are related by a deterministic, invertible function, then maximal mutual information is attained. In our case, we combine the two entropy terms and obtain mutual information between data $\mathbf{X_t}$ and prediction $\mathbf{O_t}$:
\begin{equation}
\begin{split}
\mathcal{L}_{M}&=\tfrac{1}{N}\sum_{i=1}^{N}H(\mathbf{O_t}|x^t_i)-\gamma H(\mathbf{O_t})\\
&=\tfrac{1}{N}\sum_{i=1}^{N}\sum_{j=1}^{N_C}p(o^t_{j}|x^t_i)log p(o^t_{j}|x^t_i)-\gamma \sum_{j=1}^{N_C}p(o^t_{j})logp(o^t_{j})\\
&=\sum_{i=1}^{N}\sum_{j=1}^{N_C}p(x^t_i)p(o^t_{j}|x^t_i)log p(o^t_{j}|x^t_i)-\gamma \sum_{j=1}^{N_C}p(o^t_{j})logp(o^t_{j})\\
&=H\left [\mathbf{O_t}|\mathbf{X_t}  \right ]-\gamma H\left [ \mathbf{O_t}  \right ]\approx -I(\mathbf{X_t};\mathbf{O_t}) \label{MI}
\end{split}
\end{equation}
where the first term is the entropy of conditional distribution of $\mathbf{O_t}$ which can enlarge the classifier's output of one class while suppressing those of other classes; and the second term is the entropy of marginal distribution of $\mathbf{O_t}$ which can avoid most samples being assigned to the same class. $N$ is the number of target images, and $N_C$ is the number of target categories. But without groundturth labels, how can we obtain $N_C$ and guarantee the accuracy of classifier's prediction? Benefiting from clustering-based pseudo labels, we utilize the number of clusters to substitute for $N_C$, and obtain preliminary prediction through pre-adaptation to guarantee accuracy for mutual information loss.

\subsection{Adaptation network}

As shown in Fig. \ref{fig9}, the architecture of IMAN consists of a source and target CNN, with shared weights. Maximum mean discrepancy (MMD) estimator \cite{Tzeng2014Deep,Long2015Learning,Borgwardt2006Integrating,Cafiero2006Integrating}, which is a standard distribution distance metric to measure domain discrepancy, is adopted on higher layers of network which are called adaptation layers. We simply use a fork at the top of the network after the adaptation layer. The inputs of source CNN are source labeled images while those of target CNN are target unlabeled data. The goal of training is to minimize the following loss:
\begin{equation}
\mathcal{L}=\mathcal{L}_{C}(X_{s},Y_{s})+\alpha  \sum_{l\in\mathcal{L} }MMD^2(D_s^l,D_t^l)+\beta  \mathcal{L}_{M}(X_{t})\label{eq2}
\end{equation}
where $\alpha$ and $\beta$ are the parameters for the trade-off between three terms. $\mathcal{L}_{M}(X_{t})$ is our mutual-information loss on unlabeled target data $X_{t}$. $\mathcal{L}_{C}(X_{s},Y_{s})$ denotes source classification loss on the source data $X_{s}$ and the source labels $Y_{s}$. $\mathcal{D}_*^l$ is the $l$-th layer hidden representation for the source and target examples, and $MMD^2(D_s^l,D_t^l)$ is the MMD between the source and target evaluated on the $l$-th layer representation. The empirical estimate of MMD between two domains is defined as $MM{D^2}(D_s,D_t) = \left\| {\frac{1}{M}\sum\limits_{i = 1}^M {\phi ({\rm{x}}_i^s) - } \frac{1}{N}\sum\limits_{j = 1}^N {\phi ({\rm{x}}_j^t)} } \right\|_H^2$, where $\phi$ represents the function that maps the original data to a reproducing kernel Hilbert space.

The entire procedure of IMAN is depicted in Algorithm \ref{al1}. Source classification loss supervises learning proceeds for source domain. MMD minimizes the domain discrepancy to learn domain-invariant representations. Additionally, in the pre-training stage, MMD provides more reliable underlying target representations for clustering leading to higher quality of pseudo-labels. Clustering-based pseudo-labels can improve the performance of target domain preliminarily and guarantee the accuracy of network's prediction for unsupervised MI loss. MI loss can further take full advantage of all target data, no matter whether they are successfully clustered or not, to learn larger decision margins and enhance the discrimination ability of network for target domain.

\begin{algorithm}[htb]
\caption{ Information Maximization Adaptation Network (IMAN).}
\label{al1}
\begin{algorithmic}[1]
\REQUIRE ~~\\
Source domain labeled samples $\{{x^{s}_{i}},{y^{s}_{i}}\}^{M}_{i=1}$, and target domain unlabeled samples $\{{x^{t}_{i}}\}^{N}_{i=1}$.
\ENSURE ~~\\
Network layer parameters $\Theta$.
\STATE \textbf{\emph{Stage-1: // Pre-training:}}
\STATE  Pre-train network by MMD \cite{Tzeng2014Deep} and source classification loss to minimize domain discrepancy and provide more reliable target representations for clustering;
\STATE \textbf{Repeat:}
\STATE \textbf{\emph{Stage-2: // Pre-adaptation:}}
\STATE  Adopt clustering algorithms to generate pseudo-labels of partial target images according to Eqn. (\ref{cluster}); Pre-adapt the network on them with supervision of Softmax to obtain preliminary improvement of target domain;
\STATE \textbf{\emph{Stage-3: // MI-adaptation:}}
\STATE  Adapt the network with mutual information loss according to Eqn. (\ref{eq2}) to further enhance the discrimination ability of network output;
\STATE \textbf{Until convergence}
\end{algorithmic}
\end{algorithm}

\section{Experiments on RFW}

\subsection{Racial bias experiment} \label{section}

\textbf{Experimental Settings.} We use the similar ResNet-34 architecture described in \cite{deng2018arcface}. It is trained with the guidance of Arcface loss \cite{deng2018arcface} on the CAISA-Webface \cite{yi2014learning}, and is called Arcface(CASIA) model. CASIA-Webface consists of 0.5M images of 10K celebrities in which 85\% of the photos are Caucasians. For preprocessing, we use five facial landmarks for similarity transformation, then crop and resize the faces to 112$\times$112. Each pixel ([0, 255]) in RGB images is normalized by subtracting 127.5 and then being divided by 128. We set the batch size, momentum, and weight decay as 200, 0.9 and $5e-4$, respectively. The learning rate is started from 0.1 and decreased twice with a factor of 10 when errors plateau.

\begin{figure*}
\centering
\subfigure[T-SNE]{
\label{fig4a} 
\includegraphics[width=2.3cm]{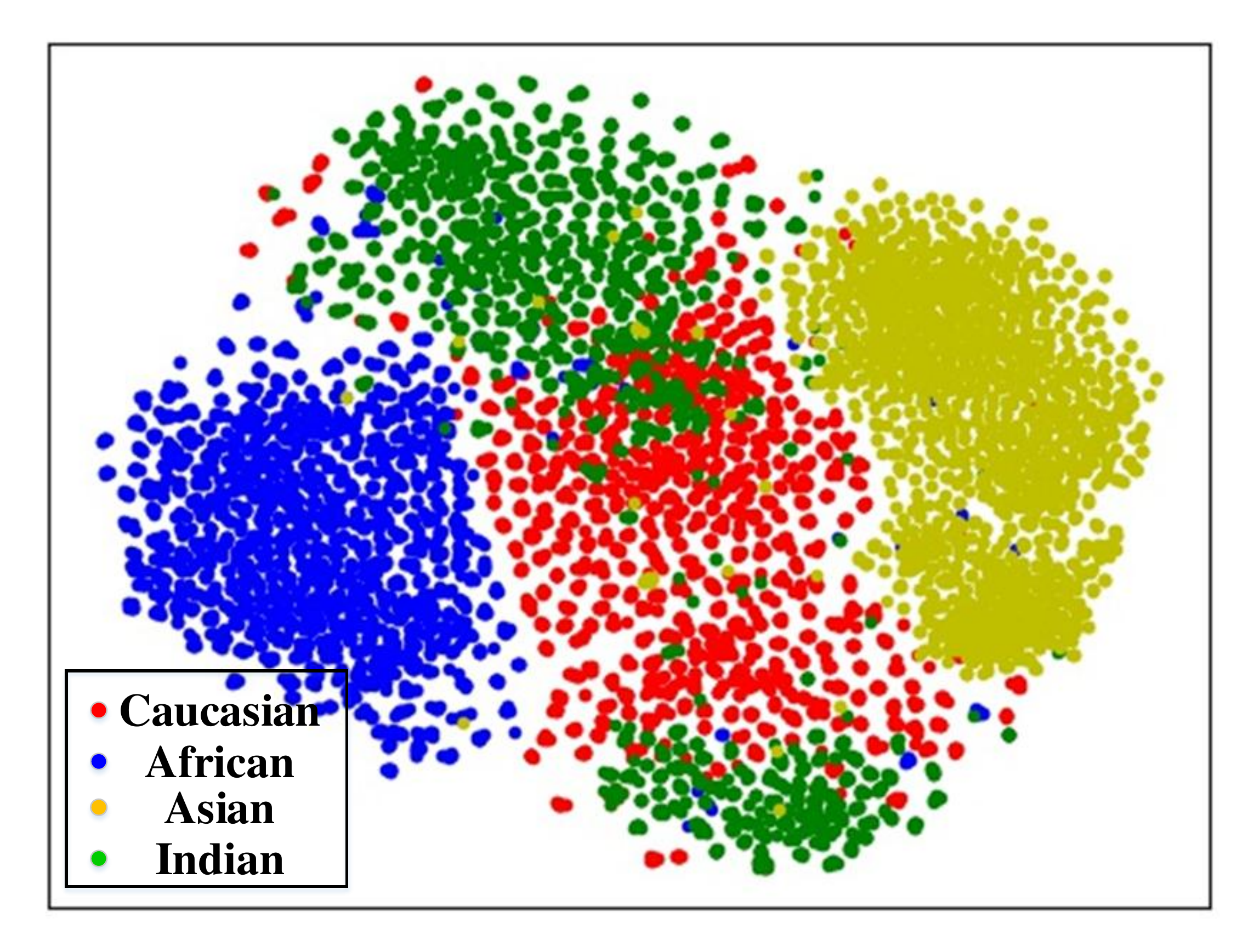}}
\hspace{0cm}
\subfigure[MMD]{
\label{fig4b} 
\includegraphics[width=2.3cm]{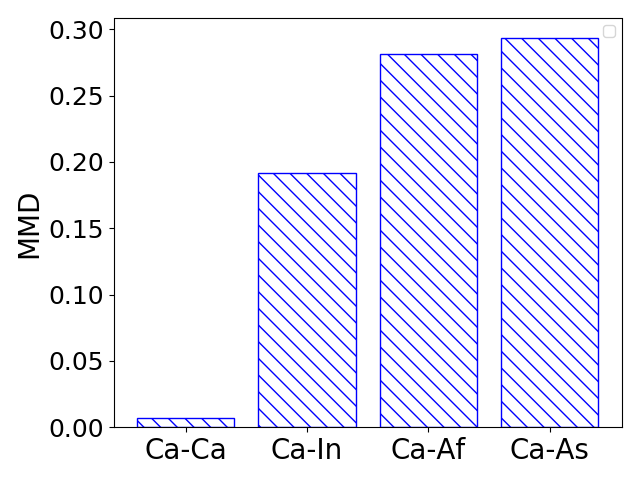}}
\hspace{0cm}
\subfigure[Distribution of cosine-distances of 6000 pairs]{
\label{fig4c} 
\includegraphics[width=12cm]{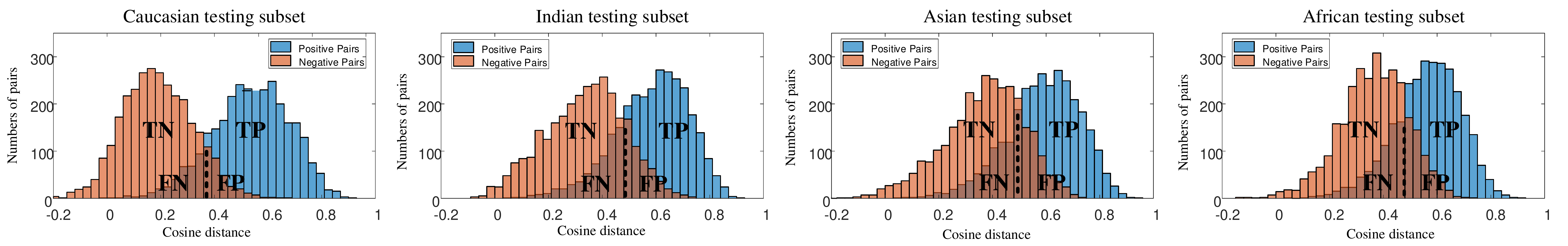}}
\caption{ (a) The feature space of four testing subsets. Each color dot represents a image belong to Caucasian, Indian, Asian or African. (b) The distribution discrepancy between Caucasians and other races measured by MMD. 'Ca', 'As', 'In' and 'Af' represent Caucasian, Asian, Indian and African, respectively. (c) Distribution of cosine-distances of 6000 pairs on Caucasian, Indian, Asian and African subset. }
\label{fig4} 
\end{figure*}

\begin{figure}
\centering
\subfigure[Center loss]{
\label{fig12a} 
\includegraphics[width=3.5cm]{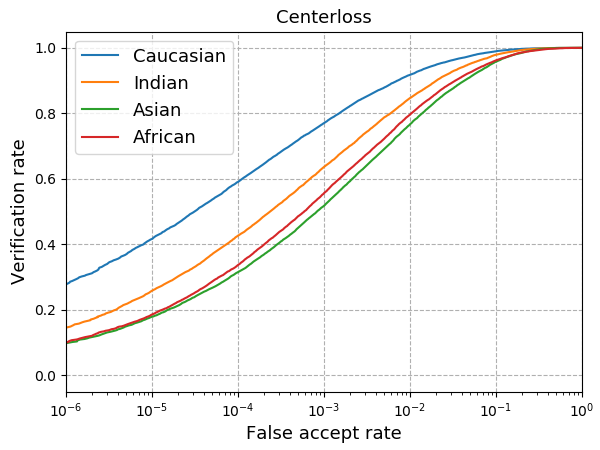}}
\hspace{0.3cm}
\subfigure[Spereface]{
\label{fig12b} 
\includegraphics[width=3.5cm]{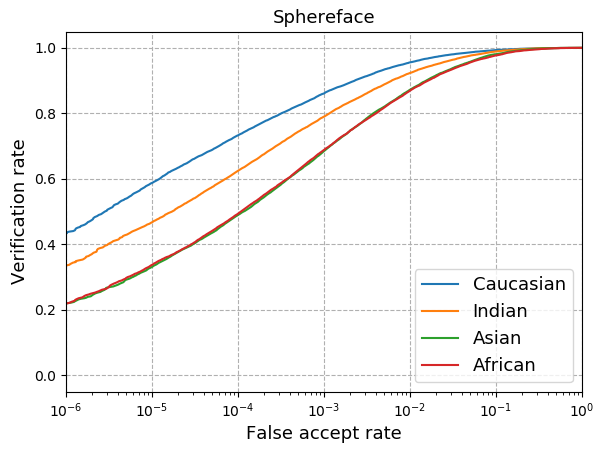}}
\subfigure[Arcface]{
\label{fig12c} 
\includegraphics[width=3.5cm]{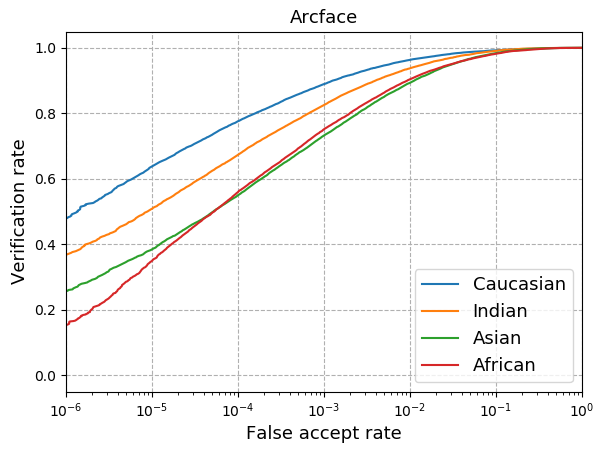}}
\hspace{0.3cm}
\subfigure[VGGFace2]{
\label{fig12d} 
\includegraphics[width=3.5cm]{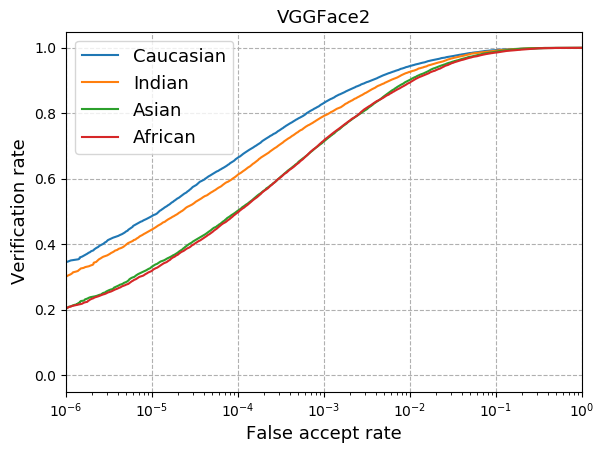}}
\caption{The ROC curves of (e) Center loss, (f) Spereface (g) Arcface, (h) VGGFace2 evaluated on all pairs.}
\label{fig5} 
\end{figure}

\textbf{Existence of racial bias.} We extract features of 6000 pairs in RFW by our Arcface(CASIA) model and compare the distribution of cosine-distances, as shown in Fig. \ref{fig4c}. The distribution of Caucasian has a more distinct margin than that of other races, which visually proves the recognition errors of non-Caucasian subjects are much higher. Then, we also examine some SOTA algorithms, i.e. Center-loss \cite{wen2016discriminative}, Sphereface \cite{liu2017sphereface}, VGGFace2 \cite{cao2017vggface2} and ArcFace \cite{deng2018arcface}, as well as four commercial recognition APIs, i.e. Face++, Baidu, Amazon, Microsoft on our RFW.
The biometric ROC curves evaluated on all pairs are presented in Fig. \ref{fig5}; the accuracies in LFW-like protocol are given in Table \ref{tab5} and its ROC curves are given in the Supplementary Material.
First, all SOTA algorithms and APIs perform the best on Caucasian testing subset, followed by Indian, and the worst on Asian and African. 
This is because that the learned representations predominantly trained on Caucasians will discard information useful for discerning non-Caucasian faces. 
Second, a phenomenon is found coincident with \cite{beveridge2008focus}: APIs which are developed by East Asian companies perform better on Asians, while APIs developed in the Western hemisphere perform better on Caucasians. 


\textbf{Existence of domain gap.} The visualization and quantitative comparisons are conducted at feature level. The deep features of 1.2K images are extracted by our Arcface(CASIA) model and are visualized respectively using t-SNE embeddings \cite{donahue2014decaf}, as shown in Fig. \ref{fig4a}. The features almost completely separate according to race. 
Moreover, we use the MMD to compute distribution discrepancy between the images of Caucasians and other races in Fig. \ref{fig4b}. From the figures, we make the same conclusions: the distribution discrepancies between Caucasians and other races are much larger than that between Caucasians themselves, which conforms that there is domain gap between races.

\textbf{Cause of racial bias.} We download more images of non-Caucasians from Website according to FreeBase celebrities \cite{freebase}, and construct an Equalizedface dataset. It contains 590K images from 14K celebrities which has the similar scale with CASIA-Webface database but is approximately race-balanced with 3.5K identities per race. Using Equalizedface as training data, we train an Arcface(Equal) model in the same way as Arcface(CASIA) model and compare their performances on 6000 difficult paris of RFW, as shown in Table \ref{tab2}. Compared with Arcface(CASIA) model, Arcface(Equal) model trained equally on all races performs much better on non-Caucasians which proves that racial bias in databases will reflect in FR algorithm. However, even with balanced training, we see that non-Caucasians still perform poorly than Caucasians. The reason may be that faces of colored skin are more difficult to extract and preprocess feature information, especially in dark situations. Moreover, we also train specific models on 7K identities of the same race, its performance is a bit lower compared to balanced training (3.5K people for each race). We believe there exists cooperative relationships among different races due to similar low-level features so that this mixture of races would improve the recognition ability.

\begin{table*}[htbp]
	\begin{center}
    \small
    \setlength{\tabcolsep}{2.5mm}{
	\begin{tabular}{c|ccc|cccc}
		\hline
         Training Databases & LFW & CFP-FP & AgeDB-30 & Caucasian & Indian & Asian & African \\ \hline \hline
         CASIA-WebFace \cite{yi2014learning} & 99.40 &\textbf{93.91} & 93.35 & 92.15 & 88.00 & 83.98 & 84.93 \\ 	\hline
         Equalizedface (ours)  & \textbf{99.55} & 92.74  &  \textbf{95.15} & \textbf{93.92}	& \textbf{92.98}	& 90.60 & \textbf{90.98} \\ \hline
         Caucasian-7000  & 99.20 & 88.00 & 94.61 & 93.68 & - & - & - \\
         Indian-7000  & 98.53 & 90.80 & 86.47 & - & 90.37  & - & - \\
         Asian-7000  & 98.05 & 87.71 & 86.05 & - & - & \textbf{91.27} & - \\
         African-7000  & 98.45 & 86.44 & 89.62 & - & - & - & 90.88   \\ \hline
	\end{tabular}}
    \end{center}
    \caption{Verification accuracy (\%) of ResNet-34 models trained with different training datasets.}
    \label{tab2}
\end{table*}

\subsection{Domain adaptation experiment}


\textbf{Datasets.} A training set with four race-subsets is also constructed according to RFW. One training subset consists of about 500K labeled images of 10k Caucasians and three other subsets contain 50K unlabeled images of non-Caucasians, respectively, as shown in Table \ref{tab4}. We use Caucasian as source domain and other races as target domains, and evaluate algorithms on 6000 pairs and all pairs of RFW.

\begin{table}[htbp]
	\begin{center}
    \small
	\begin{tabular}{c|cc|cc}
    \hline
     \multirow{2}{*}{Subsets} & \multicolumn{2}{c|}{Train} & \multicolumn{2}{c}{Test}\\
        & \# Subjects & \# Images & \# Subjects   & \# Images \\ \hline \hline
        Caucasian & 10000& 468139 & 2959 & 10196\\
        Indian &- &52285 & 2984 & 10308\\
        Asian &- & 54188 & 2492 & 9688\\
        African &- & 50588 & 2995 & 10415 \\ \hline
	\end{tabular}
    \end{center}
    \caption{Statistic of training and testing dataset.}
    \label{tab4}
\end{table}

\textbf{Implementation detail.}  
For preprocessing, we share the uniform alignment methods as Arcface(CASIA) model as mentioned above. For MMD, we follow the settings in DAN \cite{Long2015Learning}, 
and apply MMD to the last two fully-connected layers. In all experiments, we use ResNet-34 as backbone and set the batch size, momentum, and weight decay as 200, 0.9 and $5e-4$, respectively. In pre-training stage, the learning rate is started from 0.1 and decreased twice with a factor of 10 when errors plateau. In pre-adaptation stage, we pre-adapt network on pseudo-labeled target samples and source samples using learning rate of $5e-3$. In MI-adaptation stage, we adapt the network with learning rate of $1e-3$ using all source and target data. In IMAN-A(Arcface), Arcface \cite{deng2018arcface} is used as source classification loss and the parameter $\alpha$, $\beta$ and $\gamma$ are set to be 10, 5 and 0.2, respectively. In IMAN-S(Softmax), Softmax is used as source classification loss and the parameter $\alpha$, $\beta$ and $\gamma$ are set to be 2, 5 and 0.2.

\textbf{Experimental result.} Three UDA tasks are performed, namely transferring knowledge from Caucasian to Indian, Asian and African. Due to the particularity of task, very few studies have focused on UDA in FR task. The latest work is performed by Luo et al. \cite{Luo2018Adaptation} who utilizes MMD-based method, i.e. DDC \cite{Tzeng2014Deep} and DAN \cite{Long2015Learning}, to perform scene adaptation. Therefore, we also compare our IMAN with these two UDA methods. DDC adopts single-kernel MMD on the last fully-connected layers; DAN adopts multi-kernel MMD on the last two fully-connected layers.
\begin{table}[htbp]
	\begin{center}
    \small
	\begin{tabular}{c|cccc}
		\hline
         Methods & Caucasian & Indian & Asian & African \\ \hline \hline
         Softmax & 94.12 & 88.33 & 84.60 & 83.47 \\
         DDC-S \cite{Tzeng2014Deep}& - & 90.53 & 86.32 & 84.95 \\
         DAN-S \cite{Long2015Learning} & - & 89.98 &85.53 & 84.10 \\ \hline
         \textbf{IMAN-S (ours)} & - & \textbf{91.08} & \textbf{89.88}&  \textbf{89.13}\\ \hline \hline
         Arcface \cite{deng2018arcface} & 94.78 &90.48 & 86.27 & 85.13  \\
         DDC-A \cite{Tzeng2014Deep}  & - &91.63 &87.55 & 86.28 \\
         DAN-A \cite{Long2015Learning} & - & 91.78 & 87.78 & 86.30  \\ \hline
         IMAN-A (ours) & - & 93.55 & 89.87 & 88.88 \\
         \textbf{IMAN*-A (ours)} & - & \textbf{94.15} & \textbf{91.15} & \textbf{91.42} \\ \hline 
	\end{tabular}
    \end{center}
    \caption{Verification accuracy (\%) on 6000 pairs of RFW dataset. ``-S" represents the methods using Softmax as source classification loss; while ``-A" represents the ones using Arcface.}
    \label{tab3}
\end{table}

\begin{figure}
\centering
\subfigure[Indian set]{
\label{fig3a} 
\includegraphics[width=2.6cm]{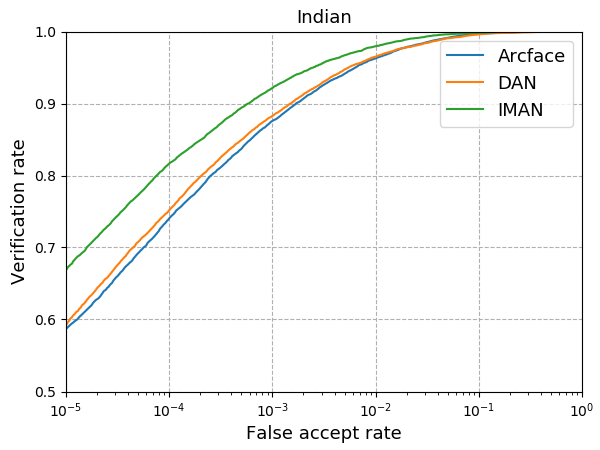}}
\hspace{0cm}
\subfigure[Asian set]{
\label{fig3b} 
\includegraphics[width=2.6cm]{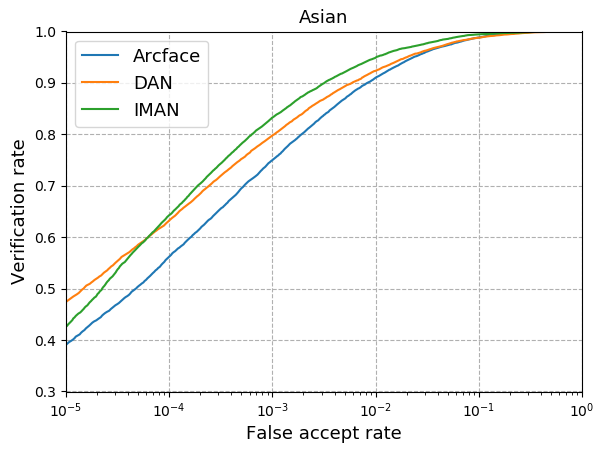}}
\subfigure[African set]{
\label{fig3c} 
\includegraphics[width=2.6cm]{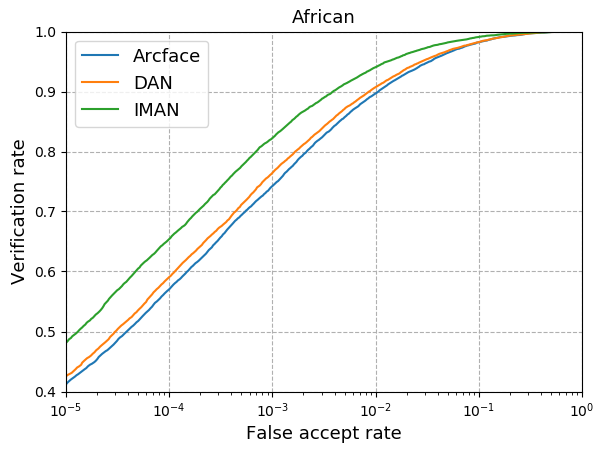}}
\caption{The ROC curves of Arcface, DAN-A, and IMAN-A models evaluated on all pairs of (a) Indian, (b) Asian and (c) African set.}
\label{fig3} 
\end{figure}

\begin{figure}[htbp]
\centering
\includegraphics[width=5.5cm]{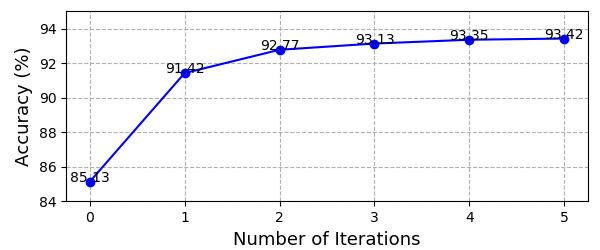}
\caption{ Verification accuracy of IMAN*-A at each iteration when performing pre-adaptation and MI-adaptation alternatively in task Caucasian$\rightarrow$African. The value at the 0-th iteration means accuracy of Arcface tested on 6K pairs of African set.}
\label{fig_iter}
\end{figure}

From Table \ref{tab3} and Fig. \ref{fig3}, we have the following observations. First, without adaptation, Arcface, which published in CVPR'19 and reported SOTA performance on the LFW and MegaFace challenges, can not obtain perfect performance on non-Caucasians due to race gap. Second, MMD-based methods, i.e. DDC and DAN, obtain limited improvement compared with Softmax and Arcface model, which confirms our thought that the popular methods by the global alignment of source and target domain are insufficient for face recognition. Third, we can find that our IMAN-S and IMAN-A both dramatically outperform all of the compared methods and IMAN-A achieves about 3\% gains over Arcface model. Furthermore, when pre-adapting network with supervision of Arcface loss instead of Softmax loss in the second stage, our IMAN-A (denoted as IMAN*-A) is further improved, and obtains the best performances with 94.15\%, 91.15\% and 91.42\% for Indian, Asian and African set. 
Especially, we further optimize IMAN*-A by performing pre-adaptation and MI-adaptation alternatively and iteratively in task Caucasian$\rightarrow$African, and show the accuracy at each iteration in Fig. \ref{fig_iter}. The performance gradually increases until convergence.

\textbf{Ablation Study.} IMAN consists of two main contributions comparing with existing UDA methods, i.e. pseudo-adaptation and MI-adaptation. To evaluate their effectiveness, we 
perform ablation study using Arcface loss as source classification loss. In Table \ref{tab6}, the results of IMAN w/o pseudo-labels are unsatisfactory because MI loss depends on pseudo-adaptation to guarantee the accuracy of classifier and only performing MI-adaptation with a randomly-initialized classifier is meaningless. To get a fair comparison, as we can see from the results of IMAN w/o MI, pseudo-adaptation is superior to baseline by about 2.3\% on average, and our IMAN outperforms pseudo-adaptation by about 1.1\% benefiting from MI-adaptation. 
It shows that each component has unique effect on reducing racial bias. 

\begin{table}[htbp]
	\begin{center}
    \small
    \setlength{\tabcolsep}{2.5mm}{
	\begin{tabular}{c|ccc}
		\hline
         Methods  & Indian & Asian & African \\ \hline \hline
         w/o pseudo-labels   &91.02 & 86.88& 85.52 \\
         w/o MI  &92.08 & 88.80 & 88.12  \\ \hline
         \textbf{IMAN-A (ours)} & \textbf{93.55} & \textbf{89.87} & \textbf{88.88} \\ \hline
	\end{tabular}}
    \end{center}
    \caption{Ablation study on 6000 pairs of RFW dataset.}
    \label{tab6}
\end{table}

\begin{figure}
\centering
\subfigure[Feature visualization]{
\label{fig20a} 
\includegraphics[width=5cm]{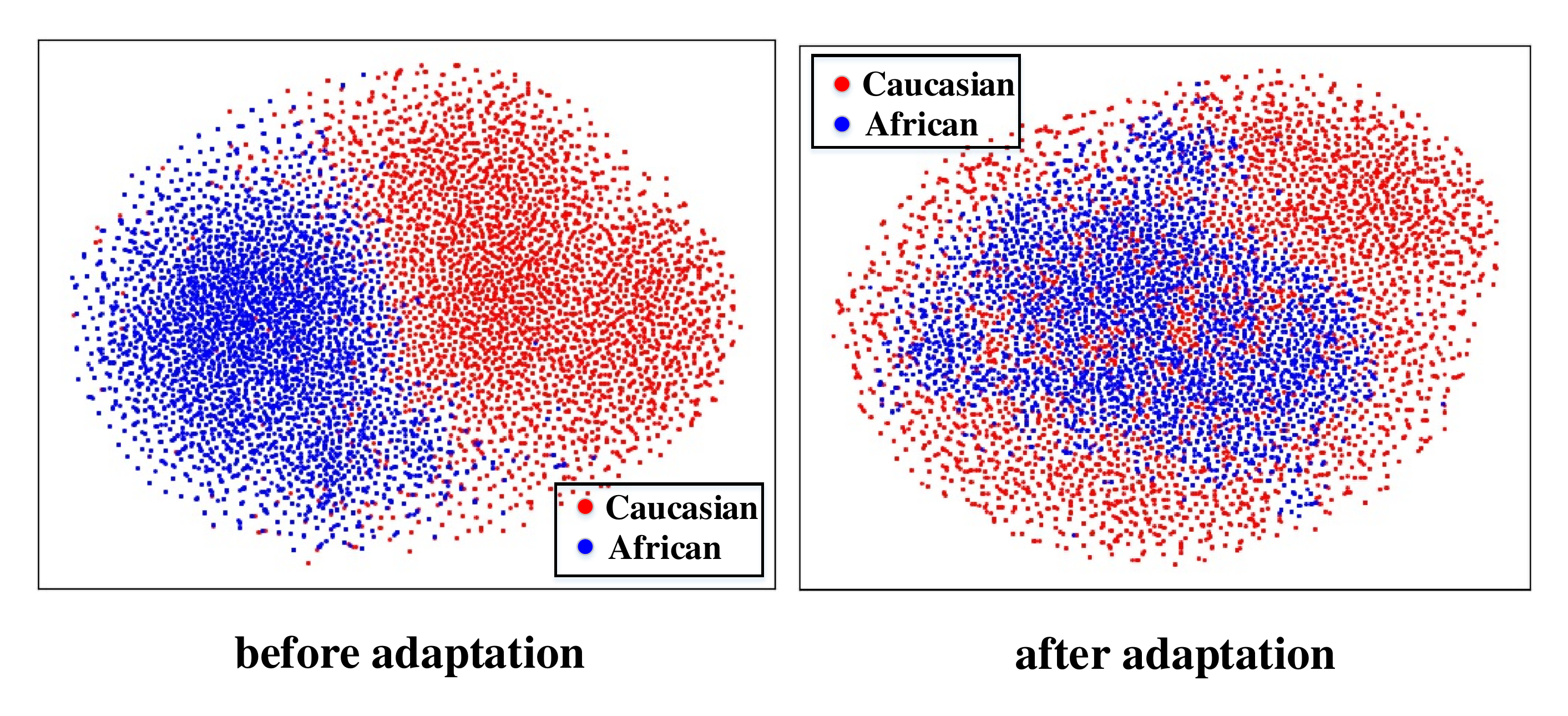}}
\hspace{0cm}
\subfigure[Domain discrepancy]{
\label{fig20b} 
\includegraphics[width=2.8cm]{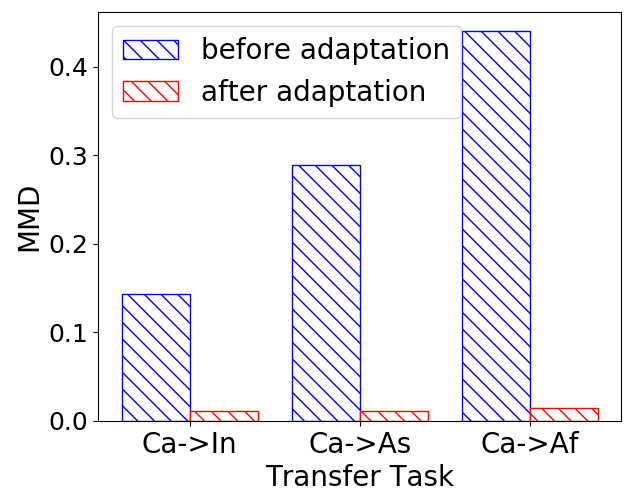}}
\caption{(a) Feature visualization in task Caucasian$\rightarrow$African. (b) Distribution discrepancy of source and target domain.}
\label{fig20} 
\end{figure}

\textbf{Visualization.} To demonstrate the transferability of the IMAN learned features, the visualization comparisons are conducted at feature level. First, we randomly extract the deep features of 10K source and target images in task Caucasian$\rightarrow$African with Arcface model and IMAN-A model, respectively. The features are visualized using t-SNE, as shown in Fig. \ref{fig20a}. After adaptation, 
more source and target data begin to mix in feature space so that there is no boundary between them. Second, we compute domain discrepancy between source and target domain using Arcface and IMAN-A activations respectively. Fig. \ref{fig20b} shows that discrepancy using IMAN-A features is much smaller than that using Arcface features.
Therefore, we conclude that our IMAN does help to minimize domain discrepancy and align feature space between two domains benefited from MMD.

\begin{table}[htbp]
\small
\begin{center}
\setlength{\tabcolsep}{3mm}{
\begin{tabular}{c|ccc}
\hline
Method & Ugly & Bad & Good\\
\hline\hline
LRPCA-face \cite{Phillips2012The} & 7.00 & 24.00 & 64.00 \\
Fusion \cite{phillips2017cross} & 15.00 & 80.00 & 98.00 \\
VGG \cite{phillips2017cross} & 26.00 & 52.00 & 85.00 \\ \hline
Arcface(CASIA) \cite{deng2018arcface} & 75.00&90.32 &96.21 \\
DAN-A \cite{Long2015Learning}&80.77 &93.66 &97.60  \\
\textbf{IMAN-A (ours)}&\textbf{85.38} &\textbf{96.00} &\textbf{98.88} \\
\hline
\end{tabular}}
\end{center}
\caption{VR at FAR of 0.001 for GBU partitions. }
\label{gbu}
\end{table}

\textbf{Additional experiments on IJB-A and GBU.} Besides race gap, there are other domain gaps which make the learnt model degenerate in target domain, e.g. different lighting condition, pose and image quality. To validate our IMAN method, we further adopt it to reduce these domain gaps by using CASIA-Webface as source domain and using GBU \cite{Phillips2012The} or IJB-A \cite{klare2015pushing} as target domain. The images in CASIA-Webface are collected from Internet under unconstrained environment and most of the figures are celebrities taken in ambient lighting. GBU is split into three partitions with face pairs of different recognition difficulty, i.e. Good, Bad and Ugly. Each partition consists of a target set and a query set, and both them contain 1085 images of 437 distinct people. The images are frontal and are taken outdoors or indoors in atriums and hallways with digital camera. IJB-A contains 5,397 images and 2,042 videos of 500 subjects, and covers large pose variations and contains many blurry video frames. The results on GBU and IJB-A databases are shown in Table \ref{gbu} and \ref{ijba}. 
After adaptation, our IMAN-A surpasses other compared methods, even better than Arcface(CASIA) model. In particular, it outperforms the SOTA counterparts by a large margin on the GBU, although it is only based on the unsupervised adaptation. 

\begin{table}
\begin{center}
\footnotesize
\begin{tabular}{c|ccc|cc}
\hline
\multirow{3}{*}{Method} & \multicolumn{3}{c|}{IJB-A: Verif.}&\multicolumn{2}{c}{\multirow{2}{*}{IJB-A: Identif.}}\\
&\multicolumn{3}{c|}{TAR@FAR's of}\\ \cline{2-6}
& 0.001 & 0.01 & 0.1 & Rank1 & Rank10\\ \hline\hline
Bilinear-CNN \cite{chowdhury2016one} & - & - & - & 58.80 & - \\
Face-Search \cite{wang2015face} & - & 73.30 & -  & 82.00 & - \\
Deep-Multipose \cite{abdalmageed2016face} & - & 78.70 & -  & 84.60 & 94.70 \\
Triplet-Similarity \cite{sankaranarayanan2016triplet} & - & 79.00 & 94.50 & 88.01 & 97.38 \\
Joint Bayesian \cite{chen2016unconstrained} & - & 83.80 & - & 90.30 & 97.70 \\
VGG \cite{parkhi2015deep} & 64.19  &  84.02  &  96.09   &  91.11  &  \textbf{98.25} \\ \hline
Arcface(CASIA) \cite{deng2018arcface} &74.19 & 87.11& 94.87& 90.68&  96.07    \\
DAN-A \cite{Long2015Learning} &80.64  & 90.87& 96.22& 92.78& 97.01 \\
\textbf{IMAN-A (ours)}& \textbf{84.19} & \textbf{91.88} & \textbf{97.05} & \textbf{94.05} & 98.04  \\ \hline
\end{tabular}
\end{center}
\caption{Verification performance (\%) of IJB-A. ``Verif" represents the 1:1 verification and ``Identif." denotes 1:N identification.}
\label{ijba}
\end{table}

\section{Conclusion}

An ultimate face recognition algorithm should perform fairly on different races. We have done the first step and create a benchmark, i.e. RFW, to fairly evaluate racial bias. Through experiments on our RFW, we first verify the existence of racial bias. 
Then, we address it in the viewpoint of domain adaptation and design a novel IMAN method to bridge the domain gap and transfer knowledge between races. The comprehensive experiments prove the potential and effectiveness of our IMAN to reduce racial bias.

{\small
\bibliographystyle{ieee}
\bibliography{egbib}
}

\end{document}